\documentclass[10pt,twocolumn,letterpaper]{article}

\usepackage[pagenumbers]{cvpr} 

\definecolor{cvprblue}{rgb}{0.21,0.49,0.74}
\usepackage[pagebackref,breaklinks,colorlinks,allcolors=cvprblue]{hyperref}

\def\paperID{*****} 
\def\confName{CVPR}
\def\confYear{2026}

\title{TK-Mamba: Marrying KAN With Mamba for Text-Driven 3D Medical Image Segmentation}

\author{Haoyu Yang\\
Zhejiang University\\
\and
Yutong Guan\\
Zhejiang University\\
\and
Meixing Shi\\
Zhejiang University\\
\and
Yuxiang Cai\\
Zhejiang University\\
\and
Jintao Chen\\
Zhejiang University\\
\and
Bing Sun\\
Research Center, National Certification Technology (Hangzhou) Co., Ltd\\
\and
Wenhui Lei\\
Shanghai Jiao Tong University\\
\and
Mianxin Liu\\
Shanghai Artificial Intelligence Laboratory\\
\and
Xiaoming Shi\\
East China Normal University\\
\and
Yankai Jiang\\
Shanghai Artificial Intelligence Laboratory\\
\and
Jianwei Yin\\
Zhejiang University\\
}

\usepackage{tabularx}
\usepackage{multirow}
\usepackage{array}

\begin{document}
\maketitle
\begin{abstract}
3D medical image segmentation is important for clinical diagnosis and treatment but faces challenges from high-dimensional data and complex spatial dependencies. Traditional single-modality networks, such as CNNs and Transformers, are often limited by computational inefficiency and constrained contextual modeling in 3D settings. To alleviate these limitations, we propose \textbf{\textit{TK-Mamba}}, a multimodal framework that fuses the linear-time Mamba with Kolmogorov-Arnold Networks (KAN) to form an efficient hybrid backbone. Our approach is characterized by two primary technical contributions. Firstly, we introduce the novel 3D-Group-Rational KAN (3D-GR-KAN), which marks the first application of KAN in 3D medical imaging, providing a superior and computationally efficient nonlinear feature transformation crucial for complex volumetric structures. Secondly, we devise a dual-branch text-driven strategy using Pubmedclip’s embeddings. This strategy significantly enhances segmentation robustness and accuracy by simultaneously capturing inter-organ semantic relationships to mitigate label inconsistencies and aligning image features with anatomical texts. By combining this advanced backbone and vision-language knowledge, \textbf{\textit{TK-Mamba}} offers a unified and scalable solution for both multi-organ and tumor segmentation. Experiments on multiple datasets demonstrate that our framework achieves state-of-the-art performance in both organ and tumor segmentation tasks, surpassing existing methods in both accuracy and efficiency. Our code is publicly available at \url{https://github.com/yhy-whu/TK-Mamba}.
\end{abstract}    
\section{Introduction}
\label{sec:Introduction}
\begin{figure}[t]
\centering
\includegraphics[width=0.48\textwidth]{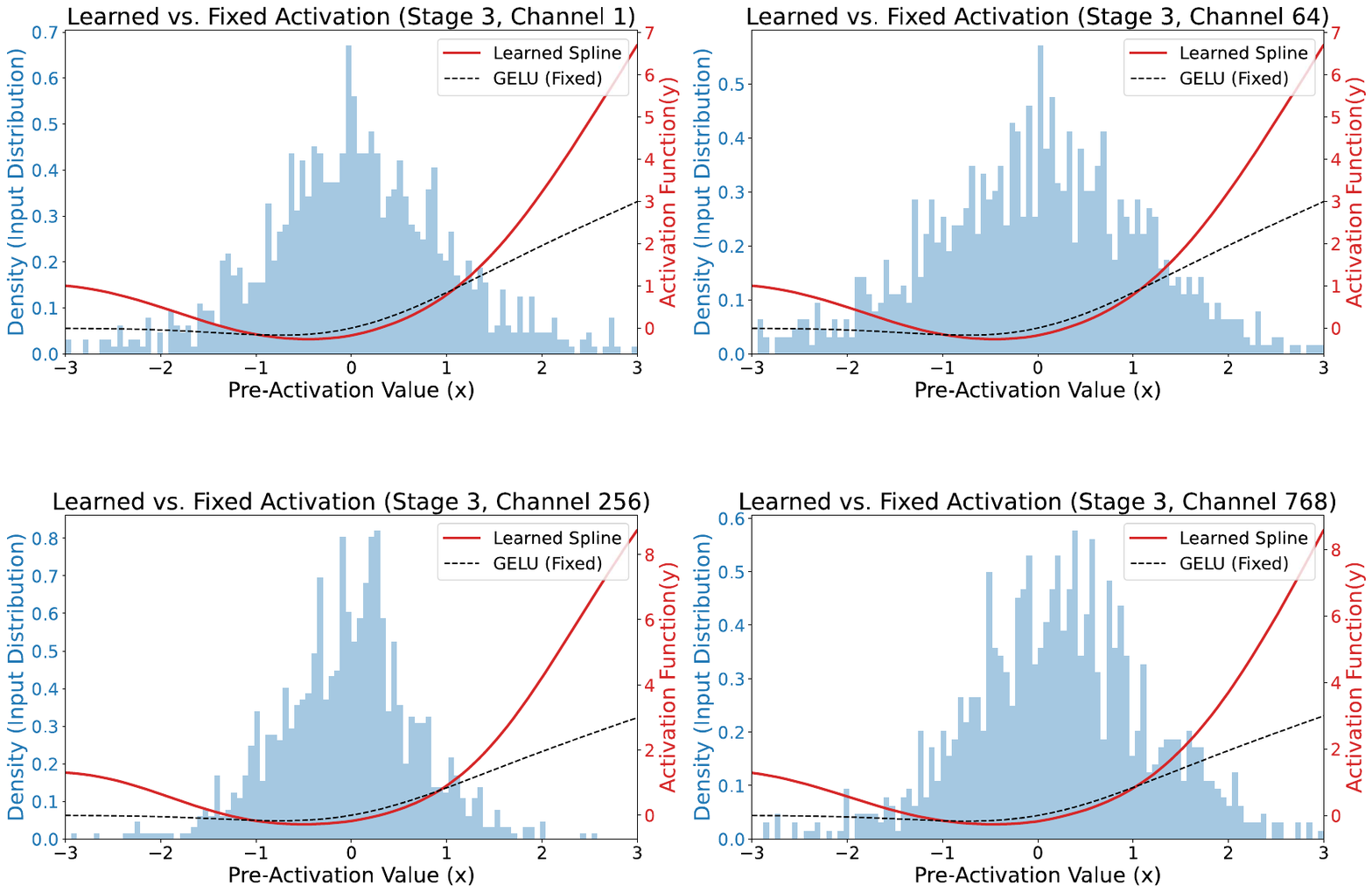}
\caption{Visualization of learned KAN splines (red) and the fixed GELU (black) for different channels in the Stage 3. 
Each function is overlaid on its corresponding pre-activation value distribution (blue). The learned splines exhibit diverse, 
data-adaptive shapes that clearly diverge from the fixed baseline.}
\label{fig:kan_visualization}
\end{figure}

3D medical image segmentation is crucial for clinical diagnosis, enabling precise delineation of anatomical and pathological structures in volumetric data such as CT and MRI scans~\cite{jiang2021ala,isensee2021nnu,ye2023uniseg,liu2023clip,chen2024transunet,zhang2024mapseg}. Multi-organ segmentation requires robust modeling of complex inter-organ semantic and spatial relationships, while tumor segmentation demands precise feature representation for specific pathological structures. Both tasks face challenges from high-dimensional data, partial annotations, and the need to capture long-range dependencies in 3D medical imaging~\cite{liu2023clip,niyas2022medical}. Traditional convolutional neural networks~\cite{yu2021convolutional,alzubaidi2021review,yuan2023effective,isensee2021nnu} 
and Transformers~\cite{chen2021transunet,zhou2023nnformer,xiao2023transformers} face challenges such as
limited receptive fields and high computational costs in 3D settings for high-resolution volumetric data~\cite{niyas2022medical,he2023transformers}.



Mamba architectures alleviate the $O(N^2)$ complexity of Transformers by leveraging an $O(N)$ selective state-space model (SSM)~\cite{gu2023mamba}. The strength of Mamba lies in its input-dependent, selective mechanism, allowing it to dynamically capture long-range dependencies based on the content of the 3D volume. However, a critical architectural mismatch arises in how these sophisticated, dynamically-aggregated features are processed. Conventional architectures feed Mamba's output into a static processing block, typically a simple MLP with a fixed activation function like GELU or SiLU. This introduces a severe expressiveness bottleneck. This one-size-fits-all non-linearity is fundamentally ill-suited for Mamba's complex output. A fixed GELU function applies the same predefined transformation indiscriminately, regardless of whether the abstract features represent subtle pathological boundaries or common anatomical structures. Consequently,The rich, nuanced information aggregated by Mamba's dynamic SSM is thus ``flattened'' or constrained by a rigid, non-adaptive functional mapping.

To resolve this bottleneck, we argue that Mamba's dynamic aggregation must be paired with an equally dynamic transformation. The advent of Kolmogorov-Arnold Networks (KANs)~\cite{liu2024kan} provides a promising solution. Instead of relying on a static activation, KANs replace the entire downstream MLP block with nodes where the activation functions themselves are parameterized, learnable B-splines. This allows the network to learn a data-adaptive and precise non-linear mapping, perfectly tailored to decipher the abstract features from Mamba. As empirically demonstrated in Figure~\ref{fig:kan_visualization}, our learned KAN splines (red) learn diverse, non-static shapes that are highly adapted to the input data distribution (blue histogram), diverging significantly from the fixed GELU baseline (black).

Despite the efficiency of Mamba and the expressiveness of KAN, a purely visual backbone remains semantically blind, it cannot differentiate between anatomically related but distinct classes, such as liver and liver tumor. To bridge this semantic gap, we draw inspiration from recent multimodal approaches that integrate visual and textual cues to enhance alignment and robustness in medical image segmentation~\cite{liu2023clip}. Accordingly, we incorporate PubMedCLIP’s textual embeddings~\cite{eslami2021does} (a medical-domain fine-tuned version of CLIP~\cite{radford2021learning}) to model inter-organ semantic relationships, mitigate label inconsistencies in partially annotated datasets, and align image features with specific anatomical descriptions of organs and tumors.

By synergistically combining Mamba’s linear-time modeling for efficient long-range dependency capture, KAN’s expressive non-linear refinement for complex anatomical structures, and PubMedCLIP-driven semantic embeddings for enhanced inter-organ relationships, \textbf{TK-Mamba} advances 3D medical image segmentation, offering a scalable solution for clinical applications. Our contributions are:
\begin{itemize}
    \item A novel \textbf{3D-GR-KAN} module tailored for 3D medical images with rational basis functions, serving as a data-adaptive non-linear refiner that replaces standard fixed-activation blocks.
    \item A \textbf{dual-branch text-driven strategy} leveraging PubMedCLIP embeddings to model inter-organ relationships and provide robust semantic priors.
    \item \textbf{State-of-the-art (SOTA)} performance on multi-organ and tumor segmentation across the MSD~\cite{antonelli2022medical} and KiTS23~\cite{heller2023kits21} datasets.
\end{itemize}
\section{Related Work}

\subsection{Sequence Modeling and Feature Representation}
Recent advances in sequence modeling and feature representation have introduced promising alternatives to traditional architectures for 3D medical image segmentation. Mamba~\cite{gu2023mamba}, a structured state-space model (SSM), achieves linear-time complexity for long-range sequence modeling, contrasting with the quadratic complexity of Transformers~\cite{gu2023mamba}. 
Building on this advancement, SegMamba~\cite{xing2024segmamba} integrates gated spatial convolution with a U-shaped architecture to fuse local and global features, achieving superior efficiency in datasets like BraTS2023~\cite{labella2023asnrmiccai}. Similarly, Tri-Plane Mamba~\cite{wang2024tri} demonstrates state-of-the-art performance in 3D CT organ segmentation. However, these Mamba-based approaches rely solely on visual features, limiting their ability to address semantic ambiguities and label inconsistencies in partially annotated datasets. 
On the other hand, Kolmogorov-Arnold Networks (KAN)~\cite{liu2024kan} offer a novel paradigm for feature representation by replacing fixed activation functions in multi-layer perceptrons with learnable edge-based activation functions. This enhances accuracy and interpretability, making it ideal for modeling complex anatomical structures in multi-organ segmentation. Yang et al.’s Group-Rational KAN (GR-KAN)~\cite{yang2024kolmogorov} further improves efficiency with rational basis functions and parameter sharing, showing promise in Transformer integration. However, KAN’s application in 3D medical imaging remains unexplored, presenting an opportunity to enhance volumetric feature representation. 
\subsection{Text-Driven Medical Image Segmentation}
Traditional 3D medical image segmentation methods primarily rely on visual data, often lacking semantic context to handle partial annotations or inter-organ relationships. Multimodal approaches that integrate visual and textual information, such as radiology reports, enhance semantic alignment and robustness to visual ambiguities~\cite{huang2020fusion}.
Liu et al.’s CLIP-Driven Universal Model~\cite{liu2023clip} uses text embeddings to capture inter-organ relationships and mitigate label inconsistencies in partially annotated datasets. Huang et al.’s dual-prompt schema~\cite{huang2024cat} employs CLIP-style cross-modal alignment to combine visual and textual prompts for robust organ and tumor segmentation. Furthermore, domain-adapted variants like PubMedCLIP~\cite{eslami2021does}, fine-tuned on PubMed medical data, improve CLIP's applicability in medical tasks by better capturing domain-specific semantics. However, these methods often rely on computationally intensive CNN or Transformer architectures, limiting scalability for 3D volumetric data.
Unlike these works, TK-Mamba synergistically integrates Mamba’s efficient sequence modeling, KAN’s expressive feature representation, and PubMedCLIP’s semantic alignment to address the computational, contextual, and semantic challenges of 3D medical image segmentation.

\section{METHODOLOGY}
\subsection{Overview of the Framework}
\label{sec:Framework}

\begin{figure*}[t]
    \centering
    \includegraphics[width=\textwidth, height=0.45\textheight, keepaspectratio]{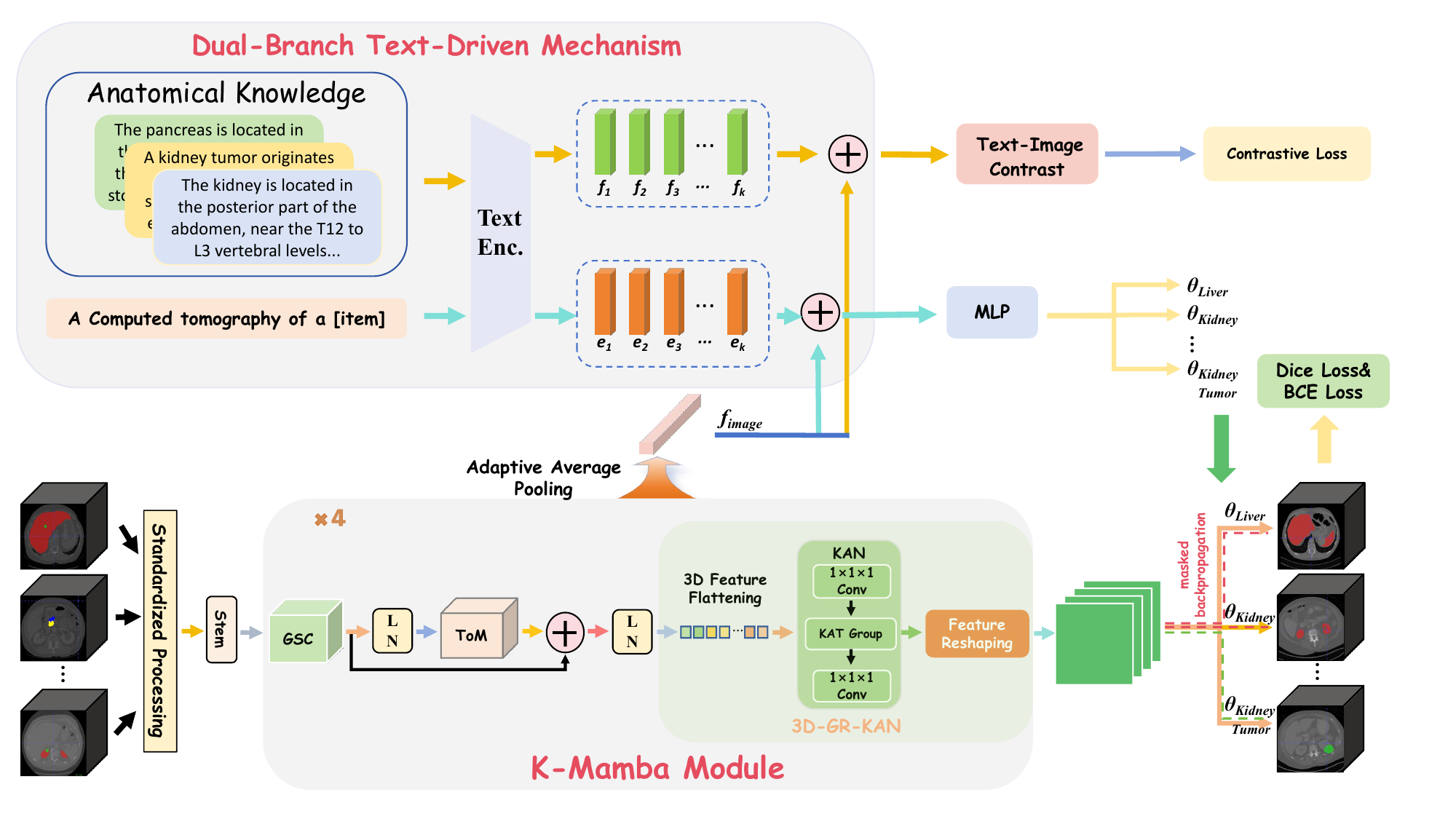}
    \caption{Overview of the TK-Mamba framework. The visual feature extraction starts with standardized preprocessing and Stem, followed by the K-Mamba Module, which includes Gated Spatial Convolution (GSC), Tri-oriented Mamba (ToM), and 3D-GR-KAN components for feature extraction. The dual-branch text-driven strategy leverages PubMedCLIP for semantic enhancement and aligns features using a text-image contrastive loss. Features are fused through adaptive average pooling and an MLP, producing the segmentation mask, supervised by a combined Dice Loss, BCE Loss, and Contrastive loss.}
    \label{fig:network-architecture}
\end{figure*}

The innovation of TK-Mamba lies in the synergy of the K-Mamba Module’s robust 3D feature extraction, alongside the dual-branch text-driven strategy, which reduces reliance on large-scale annotations and enhances segmentation accuracy for challenging structures like tumors. As illustrated in Figure~\ref{fig:network-architecture}, TK-Mamba integrates three core components: (1) a K-Mamba Module combining Gated Spatial Convolution (GSC), Tri-oriented Mamba (ToM), and 3D-GR-KAN to process 3D medical images; (2) a dual-branch text-driven strategy that leverages PubMedCLIP for semantic enhancement; and (3) a feature fusion and segmentation head that produces the final segmentation mask.

\subsection{K-Mamba Module}
\label{sec:K-Mamba Module}
The K-Mamba Module, consisting of GSC, ToM, and 3D-GR-KAN components, refines features, followed by a decoder path for feature upsampling and fusion. Following initial feature extraction by a Stem layer, the K-Mamba Module processes the resulting features $z_0 \in \mathbb{R}^{B \times 48 \times \frac{D}{2} \times \frac{H}{2} \times \frac{W}{2}}$. These features are refined by the GSC component to capture local spatial relationships, followed by the ToM and 3D-GR-KAN components to model global dependencies and enhance feature representation, respectively. The K-Mamba Module is repeated across four stages with feature dimensions $[48, 96, 192, 384]$, as depicted in Figure~\ref{fig:network-architecture}.

\subsubsection{Gated Spatial Convolution (GSC)}
The GSC module extracts local spatial features to mitigate the loss of spatial information during sequence flattening in the ToM layer. The input 3D features $z$ are processed through two paths with convolutions, each followed by instance normalization and PReLU activation, before element-wise addition and a residual connection for feature fusion. The operation is defined as:
\begin{equation}
\text{GSC}(z) = z + C_1 \big( C_3(C_3(z)) + C_1(z) \big),
\end{equation}
where \(C_k\) denotes a convolution block with kernel size \(k \times k \times k\), consisting of normalization, convolution, and PReLU activation.

\subsubsection{Mamba Module with Tri-oriented Structure (ToM)}
The Mamba module captures long-range spatial dependencies in voxel sequences of 3D medical images. Based on a state-space model (SSM), Mamba achieves linear computational complexity $O(N)$, compared to the quadratic complexity $O(N^2)$ of Transformers, making it efficient for high-resolution 3D data. The Mamba layer leverages the SSM framework, defined as:
\begin{equation}
\mathbf{h}_t = \mathbf{\overline{A}} \mathbf{h}_{t-1} + \mathbf{\overline{B}} \mathbf{x}_t, \quad \mathbf{y}_t = \mathbf{\overline{C}} \mathbf{h}_t,
\end{equation}
where $\mathbf{x}_t$ is the input sequence at timestep $t$, $\mathbf{h}_t$ is the hidden state, $\mathbf{y}_t$ is the output, and $\mathbf{\overline{A}}$, $\mathbf{\overline{B}}$, $\mathbf{\overline{C}}$ are discretized parameter matrices obtained via the Zero-Order Hold (ZOH).

To enhance 3D volumetric data modeling, we incorporate a Tri-Oriented Mamba (ToM) structure within the Mamba layer. This addresses the limitation of the original Mamba block, which models global dependencies in a single direction, by capturing feature dependencies along three directions: forward, reverse, and inter-slice. The 3D input features are flattened into three sequences along these directions, and each sequence is processed by a Mamba layer to model global information. The outputs are then fused to obtain the final 3D features:
\begin{equation}
\text{ToM}(z) = \text{Mamba}(z_f) + \text{Mamba}(z_r) + \text{Mamba}(z_s),
\end{equation}
where $z_f$, $z_r$, $z_s$ denote the flattened sequences in the forward, reverse, and inter-slice directions, respectively.

At each stage, the ToM module takes the output of the GSC module as input and processes it through the Mamba layer with the ToM structure to model global dependencies. For the $m$-th stage, the computation is defined as:
\begin{align}
\hat{z}_m &= \text{GSC}(z_m), \\
\tilde{z}_m &= \text{ToM}(\text{LN}(\hat{z}_m)) + \hat{z}_m,
\end{align}
where $\text{GSC}$ denotes the gated spatial convolution, and $\text{LN}$ is layer normalization. The output $\tilde{z}_m$ is then passed to the 3D-GR-KAN module for further feature enhancement, followed by a downsampling layer to reduce spatial resolution progressively.

\subsubsection{3D-GR-KAN Module}
\label{sec:3D-GR-KAN Module}
The 3D-GR-KAN module enhances the feature representation of the ToM module’s output through learnable nonlinear transformations, building on GR-KAN~\cite{yang2024kolmogorov}. While standard Kolmogorov-Arnold Networks (KANs)~\cite{liu2024kan} replace fixed activations with learnable B-splines for greater expressiveness, simply substituting MLPs with KANs in 3D medical imaging does not yield optimal results. This is primarily due to KANs' inefficiencies in handling high-dimensional volumetric data: the B-spline parameterization introduces significant computational overhead, as the number of spline parameters scales exponentially with input dimensionality, leading to high memory consumption and prolonged training times. For instance, in 3D medical volumes (e.g., CT scans with resolutions up to 512$\times$512$\times$512), the flattening process creates extremely long sequences, exacerbating KANs' parameter inefficiency and making them prone to overfitting or instability without extensive regularization. Moreover, KANs lack inherent mechanisms for spatial hierarchy preservation in 3D data, resulting in suboptimal capture of volumetric structures like tumor boundaries, as evidenced in benchmarks where vanilla KANs underperform in high-dimensional tasks compared to optimized variants~\cite{ta2025prkan,erdmann2024kan,yang2024kolmogorov}.

In contrast, GR-KAN improves upon KAN by incorporating rational basis functions and parameter sharing, which substantially reduce model complexity and enhance efficiency—often achieving 4$\times$ fewer parameters while maintaining or improving accuracy in high-dimensional settings~\cite{ta2025prkan}. As demonstrated in recent works like U-GRKAN~\cite{wu2025u} and MedVKAN~\cite{zhu2026medvkan}, these modifications make GR-KAN particularly suitable for medical imaging, where rational functions provide smoother approximations with lower computational cost, and parameter sharing enables cross-channel reuse to mitigate redundancy in volumetric features. We adapt GR-KAN for 3D data to leverage these advantages, ensuring effective nonlinear refinement without the bottlenecks of original KAN.

Specifically, the $l$-th 3D-GR-KAN module processes the output feature tensor of the $l$-th stage after ToM and layer normalization, denoted as $\tilde{z}_l$, with shape $[B, C_l, D_l, H_l, W_l]$, where $B$ is the batch size, $C_l$ is the feature dimension, and $D_l = \frac{D}{2^l}$, $H_l = \frac{H}{2^l}$, $W_l = \frac{W}{2^l}$ are spatial dimensions. The computation is defined as:
\begin{equation}
z_{l+1} = \text{3D-GR-KAN}(\text{LN}(\tilde{z}_l)).
\end{equation}
The processing involves three steps:
\begin{enumerate}
    \item \textbf{3D Feature Flattening}: The feature tensor is reshaped into $[B, L, C_l]$, where $L = D_l \times H_l \times W_l$, merging the 3D spatial dimensions into a sequence to enable sequence-based processing.
    \item \textbf{Nonlinear Transformation}: The flattened features are processed through a two-layer structure. First, a convolution maps the input to a hidden dimension, followed by a KAT Group activation (initialized with GELU). A second convolution maps the features back to the output dimension, with dropout applied after each activation to prevent overfitting.
    \item \textbf{Reshaping}: The transformed features are reshaped back to $[B, C_l, D_l, H_l, W_l]$, ensuring compatibility with subsequent layers while retaining 3D spatial structures.
\end{enumerate}
The 3D-GR-KAN module enhances feature representation, particularly for complex structures like tumors, by leveraging group-rational functions and dynamic reshaping, making it efficient for 3D medical images.

\subsection{Dual-Branch Text-Driven Mechanism}
\label{sec:DualBranch}
To improve the semantic relationship and recognition accuracy between categories in multi-organ segmentation tasks, we introduce a Dual-Branch Text-Driven Mechanism. This mechanism enhances the model's ability to model relationships between organs and tumors by incorporating anatomical knowledge. 

\subsubsection{Branch 1: Integration of Semantic Embeddings}
Different from traditional One-Hot encoding which represents categories as independent vectors, we adopt the PubMedCLIP text encoder~\cite{eslami2021does} (a fine-tuned version of CLIP on PubMed medical data) to convert organ names into semantic embeddings. This preserves the semantic relationships between categories and improves the model’s ability to model complex organ relationships.

\paragraph{Text Input:}The input to this method consists of the names of various organs, formatted as text prompts such as ``A Computed tomography(CT) of a [item]'', where [item] represents a specific organ category (e.g., ``liver'' or ``pancreas''). For $K$ organ categories, we construct $K$ text prompts $\{T_1, T_2, \dots, T_K\}$, where $T_k = \text{``A CT scan/MRI of a } [CLS_k]\text{''}$, and $[CLS_k]$ is the name of the $k$-th organ. We utilize the PubMedCLIP text encoder to convert each text prompt $T_k$ into a semantic embedding. For the $k$-th text prompt $T_k$, the encoding process is represented as:

\begin{equation}
e_k = \text{PubMedCLIP-Text-Encoder}(T_k),
\end{equation}
where $e_k \in \mathbb{R}^{d_{\text{c}}}$ is the semantic embedding for the $k$-th organ category, and $d_{\text{c}} = 512$ is the output dimension of the PubMedCLIP text encoder. Through this process, we obtain $K$ semantic embeddings $\{e_1, e_2, \dots, e_K\}$, forming the semantic embedding matrix $E \in \mathbb{R}^{K \times d_{\text{c}}}$, where each row $e_k$ represents the semantic embedding for the $k$-th organ category.

These semantic embeddings interact with visual features extracted from the backbone network to condition the segmentation process. Specifically, $E$ is projected to a visual-compatible dimension using a linear layer followed by ReLU activation, yielding a task encoding tensor. This tensor is then concatenated with adaptive average-pooled visual features (repeated across $K$ classes) to form conditional inputs for a controller network. The controller generates dynamic parameters (weights and biases) for a per-class segmentation head, enabling adaptive segmentation that incorporates organ-specific semantics and inter-organ relationships.

\paragraph{Semantic Relationship Modeling:}
The semantic embeddings $e_k$ capture the textual and semantic relationships between organs. The similarity is measured by the cosine similarity between the semantic embeddings:
\begin{equation}
\text{Similarity}(e_i, e_j) = \frac{e_i \cdot e_j}{\|e_i\| \|e_j\|},
\end{equation}
where $e_i$ and $e_j$ are the semantic embeddings of the $i$-th and $j$-th organs, respectively. This similarity guides the model in capturing inter-organ relationships during segmentation, providing richer prior knowledge for multi-organ segmentation tasks. Compared to traditional One-Hot encoding, the semantic embeddings generated by the PubMedCLIP text encoder offer advantages such as complex organ relationships and prior information learned during pre-training.

\subsubsection{Branch 2: Visual-Text Alignment}
\label{sec:Alignment Method}
The second branch serves to ground the model's visual features in rich semantic context. It employs a contrastive loss to align the global visual embeddings from the K-Mamba Module with a dedicated set of text embeddings derived from anatomical descriptions annotated by medical experts. This process enhances overall segmentation accuracy by enforcing semantic consistency and leveraging external anatomical knowledge.

\paragraph{Anatomical Knowledge Embedding:}
While Branch 1 utilizes semantic prompts (matrix $E$) for its dynamic segmentation head, Branch 2 requires a separate text matrix $F_t$, built from richer, more descriptive content to serve as the contrastive alignment target. 

To generate $F_t$, we leverage detailed anatomical descriptions (e.g., ``The liver is a large organ located in the upper right abdomen...'') for all $K$ classes. Following the same encoding methodology as in Branch 1, we utilize the PubMedCLIP text encoder to process each description. For each class $k$, the description is encoded into a semantic vector $f_k$. These vectors $\{f_1, \dots, f_K\}$ are then combined to form the anatomical knowledge matrix $F_t \in \mathbb{R}^{K \times d_{\text{c}}}$, which is used exclusively for this alignment task.

\paragraph{Visual Feature Extraction:}
The output features from the 4th stage, with shape $[B, C_L, D_L, H_L, W_L]$ ($C_L = 384$, $D_L = \frac{D}{2^4}$, $H_L = \frac{H}{2^4}$, $W_L = \frac{W}{2^4}$), are first mapped to a higher dimension $C_{\text{hidden}} = 768$, using a hidden convolutional layer. These features are then processed via a adaptive pooling module to produce visual embeddings $F_v \in \mathbb{R}^{B \times d_{\text{c}}}$ ($d_{\text{c}} = 512$). The adaptive pooling involves group normalization, ReLU activation, 3D adaptive average pooling to compress spatial dimensions to $(1, 1, 1)$, and a $1 \times 1 \times 1$ convolution to match the text embedding dimension:
\begin{equation}
F_v = \text{Conv3d}_{1 \times 1 \times 1}(\text{AAP}(\text{ReLU}(\text{GN}(Z')))),
\end{equation}
where $Z' \in \mathbb{R}^{B \times C_{\text{hidden}} \times D_L \times H_L \times W_L}$ is the output feature after the hidden layer, $\text{AAP}$ denotes adaptive average pooling, $\text{GN}$ denotes group normalization, and the $\text{Conv3d}_{1 \times 1 \times 1}$ maps the feature dimension to $d_{\text{c}}$.

\paragraph{Alignment and Contrastive Loss:}
Visual embeddings $F_v$ and text embeddings $F_t$ are normalized to unit vectors, resulting in $\hat{F}_v \in \mathbb{R}^{B \times d_{\text{c}}}$ and $\hat{F}_t \in \mathbb{R}^{K \times d_{\text{c}}}$. A similarity matrix $S \in \mathbb{R}^{B \times K}$ is then computed using cosine similarity:
\begin{equation}
S = \hat{F}_v \cdot \hat{F}_t^\top,
\end{equation}
where $S_{i,j}$ represents the cosine similarity between the visual embedding of the $i$-th sample and the text embedding of the $j$-th organ category.

The contrastive loss $\mathcal{L}_{\text{contrast}}$ is computed with binary cross-entropy with logits, based on ground-truth organ labels $Y \in \mathbb{R}^{B \times K}$, where $Y_{i,j} \in \{0, 1\}$ indicates the presence of the $j$-th organ in the $i$-th sample:
\begin{equation}
\mathcal{L}_{\text{contrast}} = \text{BCEWithLogitsLoss}(S, Y),
\end{equation}
where $\text{BCEWithLogitsLoss}$ combines a sigmoid activation and binary cross-entropy loss to optimize the similarity matrix $S$ against the labels $Y$. This loss encourages alignment between visual embeddings and their corresponding organ text embeddings while distinguishing them from unrelated categories. The total loss combines segmentation and contrastive losses:
\begin{equation}
\mathcal{L}_{\text{total}} = \mathcal{L}_{\text{BCE}} + \mathcal{L}_{\text{Dice}} + \mathcal{L}_{\text{contrast}},
\end{equation}
where $\mathcal{L}_{\text{BCE}}$ and $\mathcal{L}_{\text{Dice}}$ are the binary cross-entropy and Dice losses for segmentation, respectively. This alignment method facilitates semantic alignment, improving organ-specific feature understanding and introducing richer external anatomical knowledge.

\section{Experiments}
\label{sec:experiments}
\subsection{Dataset and Evaluation Metrics}

\begin{table*}[t]
\caption{Comparison of Dice and NSD scores with state-of-the-art methods for multi-organ segmentation.}
\label{tab:comparison}
\centering
{\fontsize{9}{11}\selectfont 
\begin{tabularx}{0.9\linewidth}{c|
>{\centering\arraybackslash}X>{\centering\arraybackslash}X|
>{\centering\arraybackslash}X>{\centering\arraybackslash}X|
>{\centering\arraybackslash}X>{\centering\arraybackslash}X|
>{\centering\arraybackslash}X>{\centering\arraybackslash}X|
>{\centering\arraybackslash}X>{\centering\arraybackslash}X}
\hline
\multirow{2}{*}{\textbf{Method}} & \multicolumn{2}{c|}{Liver} & \multicolumn{2}{c|}{Liver Tumor} & \multicolumn{2}{c|}{Lung Tumor} & \multicolumn{2}{c|}{Pancreas} & \multicolumn{2}{c}{Pan. Tumor} \\
\cline{2-11}
 & Dice & NSD & Dice & NSD & Dice & NSD & Dice & NSD & Dice & NSD \\
\hline
LViT & 91.38 & 86.56 & 44.82 & 47.14 & 33.78 & 29.11 & 0.39 & 3.57 & 0.24 & 1.12 \\
UNet++ & 93.71 & 86.95 & 73.55 & 84.23 & 38.58 & 45.36 & 75.20 & 78.22 & 36.17 & 38.42 \\
SegMamba & 95.82 & 92.32 & \textbf{75.25} & \textbf{87.48} & 42.94 & 45.98 & 77.72 & 79.91 & 42.49 & 45.53 \\
3D U-Net & 95.86 & 92.33 & 73.76 & 86.86 & 53.73 & 62.06 & 78.07 & 81.08 & \textbf{42.85} & 45.47 \\
Universal Model & 95.72 & 92.06 & 71.87 & 84.43 & 52.41 & 59.23 & 77.11 & 80.77 & 38.56 & 43.00 \\
TK-Mamba & \textbf{96.49} & \textbf{93.56} & 74.23 & 86.42 & \textbf{58.18} & \textbf{70.63} & \textbf{78.91} & \textbf{82.72} & 39.40 & \textbf{45.57} \\
\hline
\end{tabularx}
} 
{\fontsize{9}{11}\selectfont 
\begin{tabularx}{0.9\linewidth}{c|
>{\centering\arraybackslash}X>{\centering\arraybackslash}X|
>{\centering\arraybackslash}X>{\centering\arraybackslash}X|
>{\centering\arraybackslash}X>{\centering\arraybackslash}X|
>{\centering\arraybackslash}X>{\centering\arraybackslash}X|
>{\centering\arraybackslash}X>{\centering\arraybackslash}X|
>{\centering\arraybackslash}X>{\centering\arraybackslash}X}
\hline
\multirow{2}{*}{\textbf{Method}} & \multicolumn{2}{c|}{Hep.} & \multicolumn{2}{c|}{Hep. Tumor} & \multicolumn{2}{c|}{Colon Tumor} & \multicolumn{2}{c|}{Kidney} & \multicolumn{2}{c|}{Kidney Mass} & \multicolumn{2}{c}{Overall Avg.} \\
\cline{2-13}
 & Dice & NSD & Dice & NSD & Dice & NSD & Dice & NSD & Dice & NSD & Dice & NSD \\
\hline
LViT & 37.91 & 54.73 & 30.08 & 22.78 & 28.24 & 15.26 & \textbf{82.63} & \textbf{83.15} & 29.09 & 29.71 & 37.86 & 37.31 \\
UNet++ & 56.20 & 76.56 & 59.03 & 49.55 & 31.14 & 34.36 & 66.79 & 72.08 & 34.82 & 44.44 & 56.52 & 61.02 \\
SegMamba & 56.10 & 76.71 & \textbf{67.25} & 56.93 & 35.14 & 39.22 & 67.41 & 73.14 & 38.53 & 50.18 & 59.87 & 64.74 \\
3D U-Net & 56.51 & 76.76 & 65.59 & \textbf{57.15} & 34.14 & 39.04 & 67.20 & 73.18 & 33.52 & 46.90 & 60.12 & 66.08 \\
Universal Model & 56.50 & 76.88 & 62.74 & 54.63 & 35.57 & 40.59 & 67.33 & 73.00 & 34.30 & 45.63 & 59.21 & 65.02 \\
TK-Mamba & \textbf{56.59} & \textbf{76.93} & 63.39 & 56.84 & \textbf{38.31} & \textbf{47.49} & 67.51 & 73.09 & \textbf{38.54} & \textbf{50.22} & \textbf{61.15} & \textbf{68.35} \\
\hline
\end{tabularx}
} 
\end{table*}
\subsubsection{Single-Organ Segmentation Performance}
\begin{table}[t]
\caption{Comparison of Dice scores with state-of-the-art methods for single-organ segmentation.}
\label{tab:single_organ_results}
\setlength{\tabcolsep}{0.8mm}
\footnotesize 
\centering
\begin{tabular}{c|c|c|c|c|c|c}
\hline
\textbf{Method} & Lung T. & Panc. & Panc. T. & Hep. & Hep. T. & Avg. \\
\hline
UNETR & 55.3 & 65.7 & 37.3 & 52.3 & 53.5 & 52.8 \\
Swin-UNETR & 57.1 & 68.9 & 39.8 & 54.2 & 56.2 & 55.2 \\
Mamba-UNet & 22.6 & 61.7 & 10.4 & 49.1 & 48.7 & 38.5 \\
SegMamba & 52.2 & 77.8 & 38.1 & 57.4 & 58.5 & 56.8 \\
UNet++ & 50.5 & 77.6 & 41.2 & 57.1 & 60.4 & 57.3 \\
3D U-Net & 55.1 & 77.4 & 38.5 & 55.4 & 60.8 & 57.4 \\
Universal Model & 52.1 & 77.4 & 37.1 & 56.7 & 57.5 & 56.1 \\
nn-UNet & 59.2 & 72.3 & 40.5 & \textbf{59.9} & 65.2 & 59.4 \\
TK-Mamba & \textbf{62.6} & \textbf{78.2} & \textbf{43.9} & 57.5 & \textbf{65.4} & \textbf{61.5} \\
\hline
\end{tabular}
\end{table}

\begin{table}[t]
\caption{Model efficiency comparison, including Parameters, FLOPs, and Inference Time.}
\label{tab:model_params}
\centering
\begin{tabular}{c|ccc}
\hline
\textbf{Method} & \textbf{Params} & \textbf{FLOPs} & \textbf{Inf. Time} \\
\hline
UNet++ & 6.98 M & 563.33 G & 1.41s \\
SegMamba & 64.24 M & 655.87 G & 1.86s \\
3D U-Net & 19.07 M & 1001.80 G & 1.16s \\
Universal Model & 62.80 M & 329.59 G & 1.58s \\
nn-UNet & 88.21 M & 4248.58 G & 1.79s \\
TK-Mamba & 64.28 M & 653.07 G & 1.85s \\
\hline
\end{tabular}
\end{table}

\begin{table*}[t]
\caption{Ablation study on MSD and KiTS23 datasets.}
\label{tab:ablation}
\centering
{\fontsize{9}{11}\selectfont 
\begin{tabularx}{0.95\linewidth}{p{3cm}|>{\centering\arraybackslash}X>{\centering\arraybackslash}X|>{\centering\arraybackslash}X>{\centering\arraybackslash}X|>{\centering\arraybackslash}X>{\centering\arraybackslash}X|>{\centering\arraybackslash}X>{\centering\arraybackslash}X|>{\centering\arraybackslash}X>{\centering\arraybackslash}X}
\hline
\multirow{2}{2cm}{\centering\textbf{Method}} & \multicolumn{2}{c|}{Liver} & \multicolumn{2}{c|}{Liver Tumor} & \multicolumn{2}{c|}{Lung Tumor} & \multicolumn{2}{c|}{Pancreas} & \multicolumn{2}{c}{Pan. Tumor} \\
\cline{2-11} 
 & Dice & NSD & Dice & NSD & Dice & NSD & Dice & NSD & Dice & NSD \\
\hline
MLP+B1+B2 & 96.12 & 92.10 & 73.75 & 86.04 & 50.95 & 60.51 & 78.38 & 81.89 & 37.11 & 44.58 \\
KAN+B1+B2 & 93.95 & 88.49 & 62.22 & 74.13 & 51.25 & 60.06 & 74.24 & 77.14 & 35.42 & 40.00 \\
3D-GR-KAN & 96.04 & 92.42 & 70.75 & 82.64 & 53.08 & 62.38 & 77.86 & 81.34 & 42.50 & 45.74 \\
3D-GR-KAN+B1 & 96.22 & 92.00 & \textbf{74.56} & \textbf{87.64} & 51.66 & 62.18 & 76.91 & 80.00 & \textbf{44.31} & \textbf{52.05} \\
3D-GR-KAN+B2 & 96.43 & 92.82 & 73.11 & 86.47 & 48.64 & 58.77 & 76.82 & 80.23 & 40.25 & 45.74 \\
3D-GR-KAN+B1+B2 & \textbf{96.49} & \textbf{93.56} & 74.23 & 86.42 & \textbf{58.18} & \textbf{70.63} & \textbf{78.91} & \textbf{82.72} & 39.40 & 45.57 \\
\hline
\end{tabularx}
} 
{\fontsize{9}{11}\selectfont 
\begin{tabularx}{0.95\linewidth}{p{3cm}|>{\centering\arraybackslash}X>{\centering\arraybackslash}X|>{\centering\arraybackslash}X>{\centering\arraybackslash}X|>{\centering\arraybackslash}X>{\centering\arraybackslash}X|>{\centering\arraybackslash}X>{\centering\arraybackslash}X|>{\centering\arraybackslash}X>{\centering\arraybackslash}X|>{\centering\arraybackslash}X>{\centering\arraybackslash}X}
\hline
\multirow{2}{2cm}{\centering\textbf{Method}} & \multicolumn{2}{c|}{Hep.} & \multicolumn{2}{c|}{Hep. Tumor} & \multicolumn{2}{c|}{Colon Tumor} & \multicolumn{2}{c|}{Kidney} & \multicolumn{2}{c|}{Kidney Mass} & \multicolumn{2}{c}{Overall Avg.} \\
\cline{2-13} 
 & Dice & NSD & Dice & NSD & Dice & NSD & Dice & NSD & Dice & NSD & Dice & NSD \\
\hline
MLP+B1+B2 & 57.60 & 77.39 & 62.79 & 54.51 & 30.60 & 35.47 & 67.49 & 73.08 & 36.13 & 48.45 & 59.09 & 65.40 \\
KAN+B1+B2 & \textbf{57.97} & \textbf{77.65} & 60.61 & 51.89 & 23.13 & 26.01 & 66.76 & 71.74 & 35.93 & 47.98 & 56.15 & 61.51 \\
3D-GR-KAN & 56.89 & 77.00 & 63.55 & 55.65 & 33.73 & 37.81 & 67.39 & 73.04 & 35.95 & 47.78 & 59.77 & 65.58 \\
3D-GR-KAN+B1 & 57.04 & 77.25 & 65.09 & 55.72 & 34.49 & 39.62 & 67.25 & 72.90 & 34.53 & 46.62 & 60.21 & 66.60 \\
3D-GR-KAN+B2 & 56.89 & 77.25 & 64.62 & 55.46 & 37.03 & 44.15 & 67.36 & 73.05 & 38.00 & 49.81 & 59.92 & 66.37 \\
3D-GR-KAN+B1+B2 & 56.59 & 76.93 & 63.39 & \textbf{56.84} & \textbf{38.31} & \textbf{47.49} & \textbf{67.51} & \textbf{73.09} & \textbf{38.54} & \textbf{50.22} & \textbf{61.15} & \textbf{68.35} \\
\hline
\end{tabularx}
} 
\end{table*}
We evaluate our method on two medical imaging datasets: the Medical Segmentation Decathlon (MSD)~\cite{antonelli2022medical} and the Kidney Tumor Segmentation Challenge 2023 (KiTS23)~\cite{heller2023kits21}. The MSD dataset provides a diverse collection of 3D medical imaging tasks, and we utilize a subset of its organ and tumor structures for evaluation. The KiTS23 dataset targets segmentation of the kidney and kidney mass, where the kidney mass encompasses both tumors and cysts, presenting challenges due to their varying sizes and shapes. Each dataset is split into training and test sets at a 5:1 ratio.

We assess segmentation performance using two standard metrics: the Dice Similarity Coefficient (Dice) for volumetric overlap and the Normalized Surface Distance (NSD) for boundary accuracy within a 2 mm tolerance. Higher Dice and NSD values indicate better performance.

\subsection{Implementation Details}
All experiments were conducted using PyTorch on an NVIDIA 4090 GPU with 24GB of memory. We applied a standardized preprocessing pipeline: reorienting images to the RAS (Right-Anterior-Superior) direction, resampling to a uniform voxel spacing of 1.5 mm $\times$ 1.5 mm $\times$ 1.5 mm, intensity normalization to the range [0, 1] by scaling values from [-175, 250], and cropping to a fixed input size of $96 \times 96 \times 96$ voxels. Data augmentation included random zooming, cropping, rotations, and intensity shifts to improve generalization.

The TK-Mamba model was trained end-to-end using the AdamW optimizer~\cite{loshchilov2017decoupled} with a learning rate of $1 \times 10^{-4}$, weight decay of $1 \times 10^{-5}$, and a batch size of 1. Training ran for 2000 epochs, with a linear warmup for the first 50 epochs followed by cosine annealing schedule. For the dual-branch text-driven strategy, we used the pretrained PubMedCLIP model~\cite{eslami2021does} to generate text embeddings, which remained frozen during training. Long text descriptions were split into chunks and averaged to form final embeddings.
\begin{figure}[t!]
\centering
\includegraphics[width=0.45\textwidth]{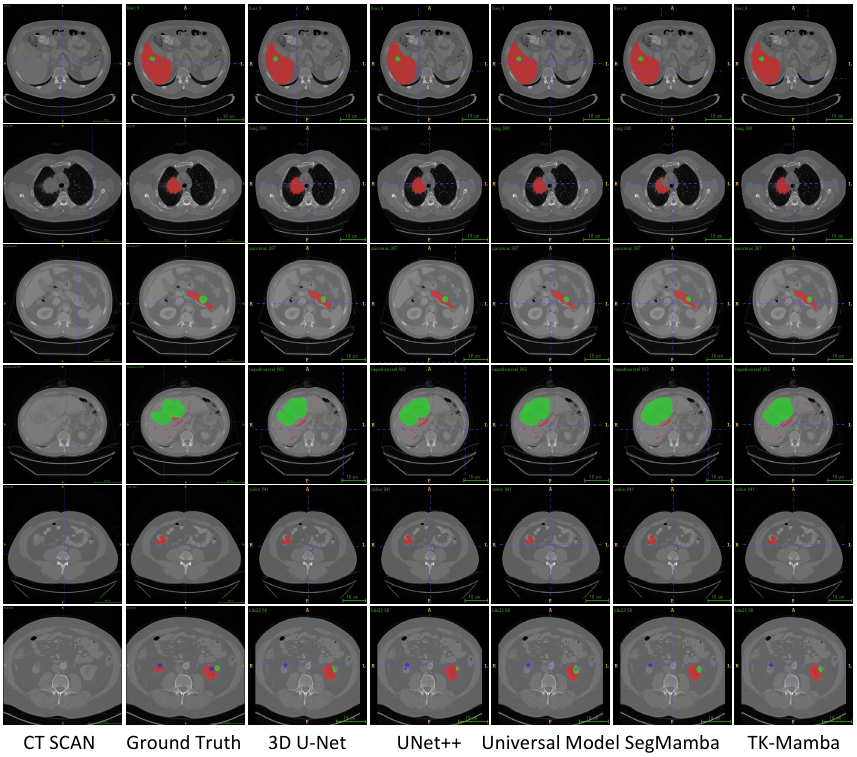}
\caption{Qualitative comparison of segmentation results on the KiTS23 and MSD datasets. Each row corresponds to a task (Liver, Lung, Pancreas, Hepatic Vessel, Colon, KiTS23).}
\label{fig:qualitative}
\end{figure}

\subsection{Comparison with SOTA Methods}
We evaluate TK-Mamba against leading 3D medical image segmentation methods. For multi-organ segmentation, we compare TK-Mamba with LViT~\cite{li2023lvit}, UNet++~\cite{zhou2018unet++}, 3D U-Net~\cite{cciccek20163d}, Universal Model~\cite{liu2023clip}, and SegMamba~\cite{xing2024segmamba}. For single-organ segmentation, we include UNETR~\cite{hatamizadeh2022unetr}, Swin UNETR~\cite{hatamizadeh2021swin}, Mamba-UNet~\cite{wang2024mamba}, SegMamba~\cite{xing2024segmamba}, UNet++~\cite{zhou2018unet++}, 3D U-Net~\cite{cciccek20163d}, Universal Model~\cite{liu2023clip}, and nn-UNet~\cite{isensee2024nnu}. Unlike most baselines, optimized for single-task settings, TK-Mamba unifies single-organ and multi-organ segmentation within a single framework, balancing performance across diverse anatomical structures. Except for LViT, which adopts its original code framework~\cite{li2023lvit}, all methods were evaluated under a unified framework with consistent preprocessing, training conditions, and evaluation protocols to ensure fair comparisons.

\subsubsection{Multi-Organ Segmentation Performance}
As shown in Table~\ref{tab:comparison}, TK-Mamba outperforms all baselines with an overall average Dice of 61.15\% and NSD of 68.35\% on MSD and KiTS23 datasets. This robust performance highlights TK-Mamba’s effectiveness in unified multi-organ segmentation, leveraging CLIP-based semantic integration and efficient 3D modeling. The superior results are attributed to its synergistic design, leveraging the dual-branch semantic integration (Section~\ref{sec:DualBranch}) and the efficient, expressive K-Mamba module (Section~\ref{sec:K-Mamba Module}).

For some complex structures, TK-Mamba remains competitive. In Hepatic Vessel segmentation (56.59\%, 76.93\%), its performance is on par with 3D U-Net and SegMamba. For challenging tumor segmentations like KiTS23 Kidney Mass (38.54\%), it matches SegMamba while surpassing other baselines. In Liver Tumors (74.23\%) and Pancreatic Tumors (39.40\%), it performs comparably to the best-performing methods. Hepatic Vessel Tumors (63.39\%, 56.84\%) are competitive with 3D U-Net (65.59\%, 57.15\%). For Kidney (67.51\%, 73.09\%), TK-Mamba trails LViT (82.63\%, 83.15\%), which performs strongly on select organs but poorly overall, notably on Pancreas (0.39\%, 3.57\%). Figure~\ref{fig:qualitative} visualizations confirm TK-Mamba’s superior accuracy and boundary precision across diverse structures.


As shown in Table~\ref{tab:single_organ_results}, TK-Mamba outperforms all baselines in single-organ segmentation, achieving an average Dice of 61.5\%, surpassing the strong nn-UNet baseline (59.4\%), 3D U-Net (57.4\%), UNet++ (57.3\%), and SegMamba (56.8\%). This robust performance underscores TK-Mamba’s effectiveness in focused segmentation. Notably, the underperformance of Mamba-UNet (38.5\%) that pairs a Mamba encoder with a standard decoder confirms our analysis from the Introduction (Section~\ref{sec:Introduction}). Mamba's linear efficiency alone is insufficient and needs to be paired with an expressive non-linear refiner, a role successfully filled by our 3D-GR-KAN module (Section~\ref{sec:3D-GR-KAN Module}).

\subsubsection{Model Efficiency}
Table~\ref{tab:model_params} summarizes model efficiency metrics, including parameters, FLOPs, and inference time. TK-Mamba balances performance and computational cost with 64.28M parameters, 653.07 GFLOPs, and an inference time of 1.85 seconds per sample. It has a similar parameter count to SegMamba (64.24M) but fewer than nn-UNet (88.21M), while its FLOPs are lower than 3D U-Net (1001.80G) and nn-UNet (4248.58G). The inference time is comparable to SegMamba (1.86 s) and slightly slower than nn-UNet (1.79 s).

\subsubsection{Ablation Study}
\label{sec:ablation}
We conduct a comprehensive ablation study, presented in Table~\ref{tab:ablation}, to dissect the individual contributions of our two primary innovations: the 3D-GR-KAN module and the Dual-Branch Text-Driven Mechanism (B1 and B2). Our full model, TK-Mamba (3D-GR-KAN+B1+B2), achieves the highest overall performance with an average Dice of 61.15\% and NSD of 68.35\%.

We validate our core design choice for the K-Mamba backbone, comparing the refiner block while keeping the text branches (B1+B2) constant. 
The data confirms our analysis from Section~\ref{sec:3D-GR-KAN Module} again:
\begin{itemize}
    \item The MLP baseline (MLP+B1+B2) achieves a solid performance of 59.09\% Dice.
    \item Naively replacing it with a standard KAN (KAN+B1+B2) causes a dramatic performance collapse to 56.15\%. This confirms that standard B-spline KANs are inefficient and unstable for high-dimensional 3D data.
    \item Our proposed 3D-GR-KAN (full model, 61.15\%) not only reverses this drop but substantially outperforms the original MLP by 2.06\% (61.15\% vs 59.09\%).
\end{itemize}
This three-way comparison proves that simply using KAN is detrimental, and our 3D-GR-KAN is the superior and essential component for unlocking expressive, high-fidelity feature representation.

The dual-branch mechanism is also proven to be highly synergistic. The backbone-only baseline, 3D-GR-KAN (59.77\% Dice), is clearly outperformed by the full model (61.15\%). The result shows that while adding Branch 1 (+B1) or Branch 2 (+B2) individually provides minor gains (60.21\% and 59.92\%, respectively), their combination is crucial for achieving the best performance. This confirms that B1 and B2 are complementary, and their synergistic application is key to robust semantic understanding.

\section{Conclusion}
\label{sec:conclusion}
We present TK-Mamba, a novel framework for multi-organ and single-organ 3D segmentation, integrating a Dual-Branch Text-Driven Mechanism (B1 and B2) with a K-Mamba Module combining Tri-oriented Mamba(ToM) and 3D-Group-Rational Kolmogorov-Arnold Networks (3D-GR-KAN). TK-Mamba delivers robust performance across MSD and KiTS23 datasets, achieving an overall multi-organ Dice of 61.15\% and NSD of 68.35\%, and a strong single-organ Dice of 61.5\%. With 64.28M parameters and 653.07 GFLOPs, TK-Mamba effectively balances segmentation accuracy with computational efficiency. Ablation studies empirically confirmed our core design choices: the 3D-GR-KAN module proved essential for solving Mamba's expressiveness bottleneck, significantly outperforming both a standard MLP and a detrimental standard KAN baseline. Furthermore, the dual branches (B1+B2) were shown to be highly synergistic, enhancing overall semantic consistency. Despite challenges with smaller structures, TK-Mamba augments clinical applications through its multimodal approach, with potential extensions to MRI and diagnostic text prompts.
{
    \small
    \bibliographystyle{ieeenat_fullname}
    \bibliography{main}
}


\end{document}